\documentclass[letterpaper, 10 pt, conference]{ieeeconf}  

\IEEEoverridecommandlockouts                              

\usepackage{amsmath,amsfonts}
\usepackage{algorithmic}
\usepackage[ruled,vlined]{algorithm2e}
\usepackage{amsmath}     
\usepackage{amssymb}     
\newcommand{\Func}[1]{\textsc{#1}}
\usepackage{array}
\usepackage[caption=false,font=normalsize,labelfont=sf,textfont=sf]{subfig}
\usepackage{textcomp}
\usepackage{stfloats}
\usepackage{url}
\usepackage{verbatim}
\usepackage{graphicx}
\usepackage{cite}
\usepackage{booktabs}
\usepackage{makecell}
\usepackage{xcolor}

\SetKwComment{TriangleComment}{$\triangleright$~}{}
\newcommand{\tctri}[1]{\TriangleComment{\footnotesize #1}}

\title{\LARGE \bf
Enhanced SIRRT*: A Structure-Aware RRT* for 2D Path Planning with Hybrid Smoothing and Bidirectional Rewiring
}

\author{Hyejeong Ryu$^{1}$
\thanks{This research was supported by the National Research Foundation of Korea grant funded by the Korea government, Ministry of Science and ICT (No. NRF-2022R1C1C1010931).}
\thanks{$^{1}$Hyejeong Ryu is with Faculty of Mechatronics Engineering,
        Kangwon National University, Gangwon 24341, South Korea
        {\tt\small hjryu@kangwon.ac.kr}}%
}

\begin{document}
\maketitle
\thispagestyle{empty}
\pagestyle{empty}
\begin{abstract}
	Sampling-based motion planners such as Rapidly-exploring Random Tree* (RRT*) and its informed variant IRRT* are widely used for optimal path planning in complex environments. However, these methods often suffer from slow convergence and high variance due to their reliance on random sampling, particularly when initial solution discovery is delayed. This paper presents Enhanced SIRRT* (E-SIRRT*), a structure-aware planner that improves upon the original SIRRT* framework by introducing two key enhancements: hybrid path smoothing and bidirectional rewiring. Hybrid path smoothing refines the initial path through spline fitting and collision-aware correction, while bidirectional rewiring locally optimizes tree connectivity around the smoothed path to improve cost propagation. Experimental results demonstrate that E-SIRRT* consistently outperforms IRRT* and SIRRT* in terms of initial path quality, convergence rate, and robustness across 100 trials. Unlike IRRT*, which exhibits high variability due to stochastic initialization, E-SIRRT* achieves repeatable and efficient performance through deterministic skeleton-based initialization and structural refinement.
\end{abstract}

\section{INTRODUCTION}
Global path planning is fundamental for autonomous mobile robot navigation, enabling robots to find collision-free paths from a start to a goal location within known environments. Among various approaches, sampling-based planners, particularly Rapidly-exploring Random Trees (RRT) and its optimal variant RRT*, have been widely adopted due to their effectiveness in handling complex obstacles and kinematic constraints.

RRT rapidly constructs a feasible path by incrementally expanding a tree from the start state toward randomly sampled states~\cite{karaman2011anytime,kuffner2000rrt}. However, a major limitation of RRT is its lack of optimality guarantees. To address this, RRT*~\cite{karaman2011sampling} connects newly sampled states to multiple neighboring nodes and performs path rewiring to minimize path cost.

Despite theoretical asymptotic optimality, RRT* often suffers from slow convergence, especially in environments containing narrow passages or large open spaces. To mitigate this, informed sampling methods have been proposed, restricting sample generation to promising regions once an initial feasible solution is found~\cite{gammell2021asymptotically}. However, these methods still depend heavily on random sampling, thus failing to fully exploit structural information inherent in the environment. Consequently, delayed initial solutions frequently lead to slower overall optimization.

Skeletonization-Informed RRT* (SIRRT*)~\cite{ryu2019improved} addresses this issue by leveraging the deterministic structure of the medial axis (skeleton) extracted from a 2D grid map. SIRRT* constructs an initial tree and path via a minimum spanning tree (MST) built upon the skeleton. This deterministic approach quickly generates reliable initial solutions with significantly reduced variance compared to random-sampling-based methods such as Informed-RRT* (IRRT*)~\cite{gammell2014informed}. Nonetheless, the initial MST-derived solutions may deviate from optimal trajectories, indicating potential for further improvement.

To overcome these limitations, we propose Enhanced SIRRT* (E-SIRRT*), which improves both initial solution quality and tree connectivity prior to informed optimization. Specifically, we introduce: (i) a hybrid path smoothing method that first generates a splined initial path via cubic interpolation and then applies collision-aware correction to produce a collision-free smoothed initial path, and (ii) a bidirectional rewiring mechanism that further improves the tree structure around this path to yield a refined initial path. These enhancements support more effective informed sampling and promote faster, more stable convergence toward high-quality solutions.

The remainder of this paper is organized as follows. Section~\ref{related_works} reviews informed sampling-based RRT planners. Section~\ref{preliminaries} briefly explains the original SIRRT* algorithm. Section~\ref{esirrt} presents the proposed E-SIRRT*, detailing the hybrid path smoothing and bidirectional rewiring procedures. Section~\ref{exp_result} provides experimental results and comparative analyses. Finally, Section~\ref{conclusion} concludes the paper and suggests directions for future research.

\section{Related Works}\label{related_works}
IRRT* enhances the convergence of RRT by sampling within an ellipsoidal informed set, defined using the current best solution cost and heuristic bounds. However, informed sampling in IRRT* commences only after an initial feasible solution is acquired, typically via uniform random sampling. This dependency can significantly delay convergence, particularly in complex environments with narrow passages.

To mitigate this limitation, IRRT*-Connect~\cite{mashayekhi2020informed} integrates informed sampling into the bidirectional RRT*-Connect framework~\cite{klemm2015rrt}. It simultaneously grows trees from the start and goal states, restricting sampling to the informed region after an initial connection is established. Subsequent improvements, including bias extension and node pruning~\cite{wang2024improved}, have further accelerated convergence and improved reliability in cluttered environments.

Recent approaches have also incorporated task-specific knowledge or learned representations into informed sampling strategies. Neural-Informed RRT~\cite{huang2024neural} leverages deep neural networks to derive informative sampling distributions from environmental point clouds. Risk-Informed RRT~\cite{chi2018risk} integrates human-centric risk metrics to promote safer navigation in human-shared environments. RBI-RRT~\cite{chen2024rbi} explicitly reconstructs and reuses previously explored tree structures to guide future expansion. AM-RRT~\cite{armstrong2021rrt} improves robustness under uncertainty by augmenting heuristic guidance with metrics such as diffusion distance.

Batch Informed Trees (BIT*)~\cite{gammell2020batch} reformulate sampling-based planning as incremental graph search over batch-sampled random geometric graphs. BIT* prioritizes edge evaluations based on estimated path costs and reuses search effort across batches. Advanced BIT* (ABIT*)~\cite{strub2020advanced} further enhances BIT* with graph-search techniques such as heuristic inflation and search truncation, accelerating initial solution discovery while preserving asymptotic optimality even in high-dimensional spaces.

SIRRT*~\cite{ryu2019improved} introduces a structure-aware variant of IRRT* that leverages environmental skeletonization to deterministically generate an initial path via an MST. This approach reduces variance in initialization and accelerates convergence compared to sampling-based methods. However, the MST-derived path is not geometrically refined, and the resulting tree structure does not support cost-efficient propagation, motivating the enhancements proposed in this paper.

To address these limitations, we propose E-SIRRT*, extending the SIRRT* framework with key improvements. Unlike the original SIRRT*, which produces suboptimal initial paths and tree connectivity, our enhanced version significantly improves initial path quality through hybrid path smoothing and refines tree connectivity using bidirectional rewiring. 

\section{Overview of SIRRT*}\label{preliminaries}
\begin{figure}[t]
	\centering
	\subfloat[]{\includegraphics[width=1.0\linewidth]{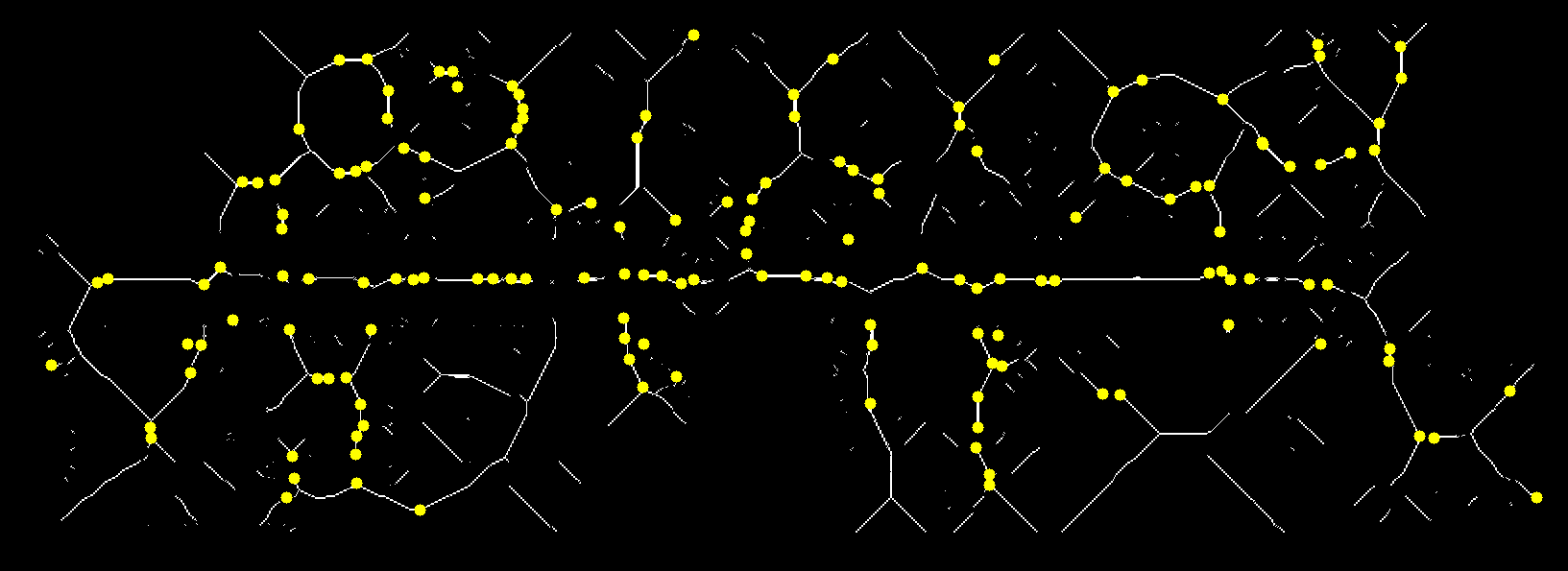}
		\label{fig:ske_sirrt}}
	\hfil
	\subfloat[]{\includegraphics[width=1.0\linewidth]{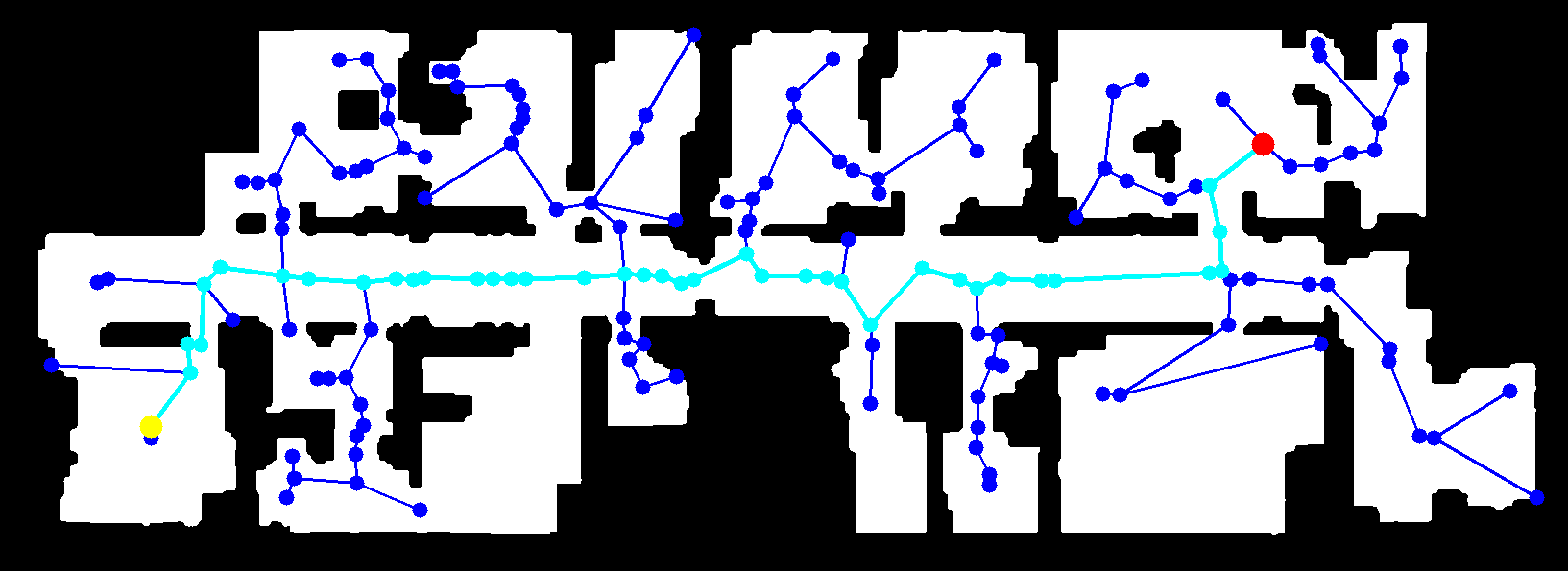}
		\label{fig:initial_sirrt}}
	\hfil
	\subfloat[]{\includegraphics[width=1.0\linewidth]{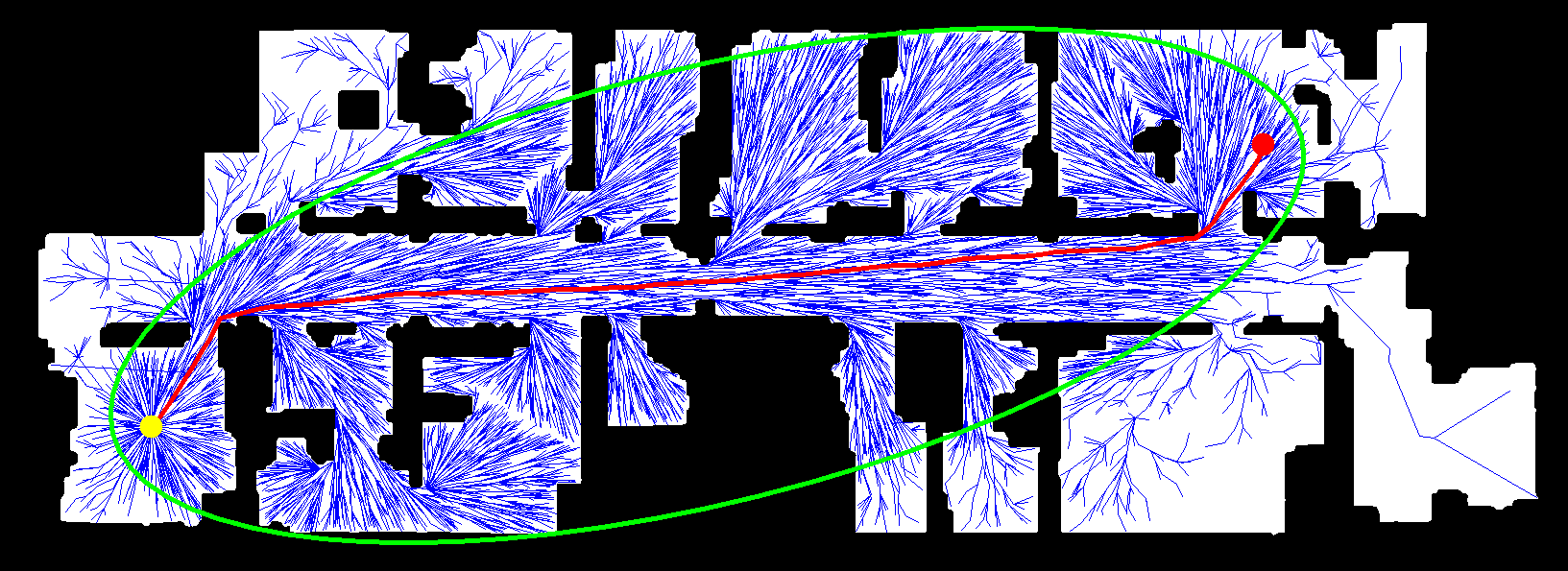}
		\label{fig:final_sirrt}}	
	\caption{Overview of the SIRRT* algorithm:
		(a) Skeletonization of the binary occupancy grid map via morphological thinning; Harris corner nodes are marked in yellow.
		(b) Initial tree (blue nodes and lines) and initial path (cyan line) constructed using an MST over the extracted corner nodes, start (yellow dot), and goal (red dot) positions.
		(c) Optimization phase via informed sampling within an ellipsoidal region (green ellipse); the final optimized path is shown in red.}		
	\label{fig:sirrt}
\end{figure}

The SIRRT* algorithm~\cite{ryu2019improved} leverages deterministic structural information extracted from a 2D grid map to efficiently generate initial solutions, addressing the limitations of purely random sampling methods. It begins by computing the medial axis (skeleton) of the free space using morphological thinning, which effectively captures the topological structure of the environment. Harris corner detection is then applied to the skeleton to identify salient points that represent meaningful structural features (Fig.~\ref{fig:ske_sirrt}).

Using these skeleton-derived nodes, along with the start and goal positions, an MST is constructed via Prim's algorithm~\cite{cormen2009introduction}. This tree captures the connectivity of the environment, and an initial path is extracted by tracing the MST from the goal node back to the start node (Fig.~\ref{fig:initial_sirrt}). This deterministic initialization significantly reduces computation time and solution variance compared to stochastic methods such as IRRT*~\cite{gammell2014informed}. Furthermore, the initial solution defines a focused ellipsoidal sampling region for subsequent informed optimization (Fig.~\ref{fig:final_sirrt}).

However, the MST-derived path often contains unnatural turns and geometric irregularities, which may hinder efficient cost propagation during the optimization phase. These limitations motivate the enhancements proposed in the next section.

\section{Proposed Method: Enhanced SIRRT*}\label{esirrt}
E-SIRRT* introduces two improvements to the original SIRRT* algorithm: hybrid path smoothing and bidirectional rewiring. These improvements are applied prior to informed optimization and support more accurate cost propagation and faster, more stable convergence.

\subsection{Initial Path Enhancement by Hybrid Path Smoothing} \label{hybrid_path_smoothing}
Although the MST-derived path connects the start and goal nodes deterministically and efficiently, it often lacks geometric smoothness due to the sparse and piecewise nature of the skeleton and the greedy structure of MST construction. Such irregularities can reduce the path?s suitability for real-world execution, where curvature continuity is often required, and impede optimization convergence.

To address this, we apply a hybrid path smoothing procedure to refine the MST-derived path. This process consists of two stages: (i) spline fitting to generate a splined initial path with improved geometric continuity, and (ii) collision-aware correction to ensure feasibility, yielding a smoothed initial path. The full procedure is summarized in Algorithm~\ref{alg:hybrid_path_smoothing}.

\subsubsection{Hybrid Path Smoothing}
In the first stage, the initial MST path $\mathcal{P}_{\text{init}}$ is sparsely subsampled using a fixed interval $d$ to eliminate unnecessary waypoints (line~2). The resulting control point sequence is denoted by $\mathcal{P}_{\text{sub}} = \{\mathbf{p}_i\}_{i=0}^{m}$, where each point $\mathbf{p}_i = [x_i, y_i]^T$, and the normalized parameter values are given by $u_i = i/m$, with $m = |\mathcal{P}_{\text{sub}}| - 1$ denoting the number of intervals between subsampled points.

Next, we fit two independent cubic spline functions to the $x$ and $y$ coordinate sequences over the domain $[0, 1]$ using the subroutine \textsc{CubicSplineFit} (lines~4--5; see Algorithm~\ref{alg:cubic_spline_fit}). This routine computes the spline coefficients $a_i$, $b_i$, $c_i$, and $d_i$ for each segment over $[u_i, u_{i+1}]$, such that
\begin{equation}\label{eq_spline}
	s_i(u) = a_i + b_i(u - u_i) + c_i(u - u_i)^2 + d_i(u - u_i)^3,
\end{equation}
for $i = 0, \dots, m-1$.

Once the spline coefficients are obtained, the path is evaluated at $N + 1$ uniformly spaced parameter values $u_k = \frac{k}{N}$, for $k = 0, \dots, N$ (lines~6--13). Each $u_k$ is used to compute interpolated coordinates $[x(u_k), y(u_k)]$ within the corresponding segment. This yields a densely sampled splined initial path $\mathcal{P}_{\text{spline}}$ that smoothly interpolates the original control points.

In the second stage (line~14), the spline-fitted path is validated for collision. Although geometrically smooth, it may contain segments that are infeasible due to obstacles. To ensure feasibility, we apply the subroutine \textsc{CollisionAwareCorrection} (Algorithm~\ref{alg:collision_correction}), which replaces invalid segments with safe alternatives drawn from the original MST-derived path. The result is a collision-free smoothed initial path $\mathcal{P}_{\text{smooth}}$, which is used to update the initial tree structure in the next phase.

\begin{algorithm}[t]
	\caption{Hybrid Path Smoothing}
	\label{alg:hybrid_path_smoothing}
	\KwIn{Initial path $\mathcal{P}_{\text{init}}$, subsampling interval $d$, number of interpolation intervals $N$, map image $I_{\text{map}}$}
	
	\KwOut{Collision-aware smoothed path $\mathcal{P}_{\text{smooth}}$}
	
	\SetKwProg{Fn}{function}{:}{}
	\Fn{\Func{HybridPathSmoothing}($\mathcal{P}_{\text{init}}, d, N, I_{\text{map}}$)}{
		
		\tctri{Subsample the path by distance $d$}
		$\mathcal{P}_{\text{sub}} \leftarrow \Func{SubsamplePath}(\mathcal{P}_{\text{init}}, d)$\;
		$m \leftarrow |\mathcal{P}_{\text{sub}}| - 1$\;
		\tctri{Let $(x_i, y_i)$ be the coordinates of $\mathcal{P}_{\textrm{sub}}[i]$, and $u_i = i/m$}
		\tctri{Fit cubic spline coefficients}
		$\{a_i^x, b_i^x, c_i^x, d_i^x\}_{i=0}^{m-1} \leftarrow \Func{CubicSplineFit}([x_0, \dots, x_m], [u_0, \dots, u_m])$\;
		$\{a_i^y, b_i^y, c_i^y, d_i^y\}_{i=0}^{m-1} \leftarrow \Func{CubicSplineFit}([y_0, \dots, y_m], [u_0, \dots, u_m])$\;
		
		\tctri{Evaluate spline at $N+1$ uniform samples}
		$\mathcal{P}_{\text{spline}} \leftarrow [\, ]$\;
		\For{$k \leftarrow 0$ \KwTo $N$}{
			$u_k \leftarrow \frac{k}{N}$\;
			Find segment index $i$ such that $u_k \in [u_i, u_{i+1}]$\;
			$\delta \leftarrow u_k - u_i$\;
			$x_k \leftarrow a_i^x + b_i^x \delta + c_i^x \delta^2 + d_i^x \delta^3$\;
			$y_k \leftarrow a_i^y + b_i^y \delta + c_i^y \delta^2 + d_i^y \delta^3$\;
			$\mathcal{P}_{\text{spline}}.\Func{append}([x_k, y_k])$\;
		}
		
		\tctri{Apply collision-aware correction}
		$\mathcal{P}_{\text{smooth}} \leftarrow \Func{CollisionAwareCorrection}(\mathcal{P}_{\text{spline}}, \mathcal{P}_{\text{init}}, I_{\text{map}})$\;
		
		\Return{$\mathcal{P}_{\text{smooth}}$}\;
	}
\end{algorithm}

\begin{algorithm}[t]
	\caption{Cubic Spline Fit}
	\label{alg:cubic_spline_fit}
	\KwIn{Data points $[z_0, z_1, \dots, z_m]$, knots $[u_0, u_1, \dots, u_m]$}
	\KwOut{Spline coefficients $\{a_i, b_i, c_i, d_i\}_{i=0}^{m-1}$ for segments $[u_i, u_{i+1}]$}
	
	\SetKwProg{Fn}{function}{:}{}
	\Fn{\Func{CubicSplineFit}($[z_0, \dots, z_m], [u_0, \dots, u_m]$)}{
		
		\tctri{Assume uniform spacing}
		$h \leftarrow u_{i+1} - u_i$\;
		
		\tctri{Construct the right-hand side vector $\mathbf{b}$}
		\For{$i \leftarrow 1$ \KwTo $m-1$}{
			$b_i \leftarrow 3 \left( \frac{z_{i+1} - z_i}{h} - \frac{z_i - z_{i-1}}{h} \right)$\;
		}
		
		\tctri{Apply natural boundary conditions}
		$M_0 \leftarrow 0$, $M_m \leftarrow 0$\;
		
		\tctri{Solve tridiagonal system $A \mathbf{M} = \mathbf{b}$}
		Compute $M_i = s_i''(u_i)$ for $i = 1, \dots, m-1$\;	
		
		\tctri{Compute spline coefficients on each interval $[u_i, u_{i+1}]$}
		\For{$i \leftarrow 0$ \KwTo $m-1$}{
			$a_i \leftarrow z_i$\;
			$b_i \leftarrow \frac{z_{i+1} - z_i}{h} - \frac{h}{3}(2M_i + M_{i+1})$\;
			$c_i \leftarrow M_i$\;
			$d_i \leftarrow \frac{M_{i+1} - M_i}{3h}$\;
		}
		
		\Return{$\{a_i, b_i, c_i, d_i\}_{i=0}^{m-1}$}\;
	}
\end{algorithm}

\subsubsection{Cubic Spline Fit}
The subroutine \textsc{CubicSplineFit} (Algorithm~\ref{alg:cubic_spline_fit}) computes the coefficients of a natural cubic spline that interpolates a sequence of scalar values $\{z_0, z_1, \dots, z_m\}$ over a uniformly spaced parameter domain $\{u_0, u_1, \dots, u_m\}$. The objective is to compute segment-wise spline coefficients, as defined in~(\ref{eq_spline}), to ensure smooth interpolation across all subintervals $[u_i, u_{i+1}]$.

Assuming uniform spacing $h = u_{i+1} - u_i$ (line~2), the method constructs a tridiagonal linear system $A \mathbf{M} = \mathbf{b}$ to solve for the second derivatives $\mathbf{M} = [M_0, \dots, M_m]^T$ at the knots (lines~3--4). Natural boundary conditions are imposed by setting $M_0 = M_m = 0$, enforcing zero curvature at the endpoints.

The matrix $A$ is symmetric and tridiagonal, with size $(m{-}1) \times (m{-}1)$, derived from continuity constraints on the second derivatives of adjacent spline segments. Under uniform spacing, its structure is given by
\begin{equation}
	A = \frac{1}{h}
	\begin{bmatrix}
		4 & 1 & 0 & \cdots & 0 \\
		1 & 4 & 1 & \cdots & 0 \\
		0 & 1 & 4 & \ddots & \vdots \\
		\vdots & \ddots & \ddots & \ddots & 1 \\
		0 & \cdots & 0 & 1 & 4
	\end{bmatrix},
\end{equation}
where each row enforces a second-derivative continuity constraint at an interior knot. The first and last rows are excluded due to the imposed boundary conditions. This system can be efficiently solved using the Thomas algorithm~\cite{quarteroni2007scientific}.

Once the second derivatives are obtained, the spline coefficients $\{a_i, b_i, c_i, d_i\}_{i=0}^{m-1}$ for each segment are computed analytically (lines~7--11). These coefficients ensure $\mathcal{C}^2$ continuity across the domain and define a smooth, twice-differentiable interpolant over $[0, 1]$.

\begin{algorithm}[]
	\caption{Collision-Aware Correction}
	\label{alg:collision_correction}
	\KwIn{Spline path $\mathcal{P}_{\text{spline}}$, fallback path $\mathcal{P}_{\text{init}}$, map image $I_{\text{map}}$}
	\KwOut{Collision-safe smoothed path $\mathcal{P}_{\text{smooth}}$}
	
	\SetKwProg{Fn}{function}{:}{}
	\Fn{\Func{CollisionAwareCorrection}($\mathcal{P}_{\text{spline}}, \mathcal{P}_{\text{init}}, I_{\text{map}}$)}{
		
		\tctri{Initialize with the first point}
		$\mathcal{P}_{\text{smooth}} \leftarrow [\mathcal{P}_{\text{spline}}[0]]$\;
		$prev \leftarrow \mathcal{P}_{\text{spline}}[0]$\;
		
		\tctri{Iterate through remaining spline points}
		\For{$i \leftarrow 1$ \KwTo $|\mathcal{P}_{\text{spline}}| - 1$}{
			$curr \leftarrow \mathcal{P}_{\text{spline}}[i]$\;
			
			\If{$\Func{ObstacleFree}(prev, curr, I_{\text{map}})$}{
				$\mathcal{P}_{\text{smooth}}.\Func{append}(curr)$\;
				$prev \leftarrow curr$\;
			}
			\Else{
				\tctri{Fallback: nearest feasible point from $\mathcal{P}_{\text{init}}$}
				$best \leftarrow \text{None}, \; bestDist \leftarrow \infty$\;
				
				\ForEach{$f \in \mathcal{P}_{\text{init}}$}{
					\If{$\Func{ObstacleFree}(prev, f, I_{\text{map}})$}{
						\If{$\|f - curr\| < bestDist$}{
							$best \leftarrow f$, $bestDist \leftarrow \|f - curr\|$\;
						}
					}
				}
				
				\If{$best \ne \text{None}$ and $best \ne prev$}{
					$\mathcal{P}_{\text{smooth}}.\Func{append}(best)$\;
					$prev \leftarrow best$\;
				}
			}
		}
		
		\Return{$\mathcal{P}_{\text{smooth}}$}\;
	}
\end{algorithm}

\subsubsection{Collision-Aware Correction}
The subroutine \textsc{CollisionAwareCorrection} (Algorithm~\ref{alg:collision_correction}) refines the splined initial path to ensure that all segments are collision-free. While spline interpolation improves geometric continuity, it does not account for obstacles and may generate infeasible segments. This subroutine performs segment-wise validation and applies fallback corrections using points from the original MST-derived path $\mathcal{P}_{\text{init}}$.

The corrected path $\mathcal{P}_{\text{smooth}}$ is initialized with the first point of $\mathcal{P}_{\text{spline}}$ (lines~2--3). The algorithm then iterates over the remaining spline points (lines~4--9), checking whether the segment from the last valid point to the current point is obstacle-free. If the segment is valid, the current point is appended to $\mathcal{P}_{\text{smooth}}$ (line~7); otherwise, a fallback procedure is triggered (lines~10--17).

During fallback, the algorithm searches for the nearest point $f \in \mathcal{P}_{\text{init}}$ that forms a collision-free segment with the last valid point. Among all feasible candidates, the one with the smallest Euclidean distance to the current spline point is selected and appended (lines~13--16), provided it differs from the previous point (line~17). Finally, the function returns the collision-free smoothed initial path $\mathcal{P}_{\text{smooth}}$ (line~18), which preserves the geometry of the spline wherever feasible and applies minimal corrections only when necessary.

\begin{algorithm}[t]
	\caption{Bidirectional Rewiring}
	\label{alg:bidirectional_rewiring}
	\KwIn{Tree $\mathcal{T}$, Smoothed path $\mathcal{P}_{\text{smooth}}$}
	\KwOut{Rewired tree $\mathcal{T}$}
	
	\SetKwProg{Fn}{function}{:}{}
	\Fn{\Func{BidirectionalRewiring}($\mathcal{T}, \mathcal{P}_{\text{smooth}}$)}{
		\ForEach{$p \in \mathcal{P}_{\text{smooth}}$}{
			$N_{\text{near}} \leftarrow \Func{FindNeighbors}(p, \mathcal{T})$\;
			
			\tctri{Forward rewiring}
			\ForEach{$q \in N_{\text{near}}$}{
				\If{$\Func{Cost}(p) + \Func{Cost}(p, q) < \Func{Cost}(q)$ \textbf{and} $\Func{ObstacleFree}(p, q)$}{
					$\Func{Parent}(q) \leftarrow p$\;
					\Func{UpdateCost}($q$)\;
				}
			}
			
			\tctri{Reverse rewiring}
			\ForEach{$q \in N_{\text{near}}$}{
				\If{$\Func{Cost}(q) + \Func{Cost}(q, p) < \Func{Cost}(p)$ \textbf{and} $\Func{ObstacleFree}(q, p)$}{
					$\Func{Parent}(p) \leftarrow q$\;
					\Func{UpdateCost}($p$)\;
				}
			}
		}
		\Return{$\mathcal{T}$}\;
	}
\end{algorithm}

\subsection{Initial Tree Refinement by Bidirectional Rewiring} \label{sec:bidirectional_rewiring}
Although the smoothed initial path $\mathcal{P}_{\text{smooth}}$ improves geometric quality by eliminating jagged turns, the initial tree $\mathcal{T}$ constructed from the MST does not reflect this improvement in either connectivity or cost structure. As a result, the tree may be misaligned with the smoothed path and may not support effective cost propagation. To resolve this, we refine the tree around $\mathcal{P}_{\text{smooth}}$ using a bidirectional rewiring strategy, which updates parent-child relationships and path costs based on proximity and feasibility. The full procedure is summarized in Algorithm~\ref{alg:bidirectional_rewiring}.

\begin{figure}[t]
	\centering
	\subfloat[]{\includegraphics[width=0.48\linewidth]{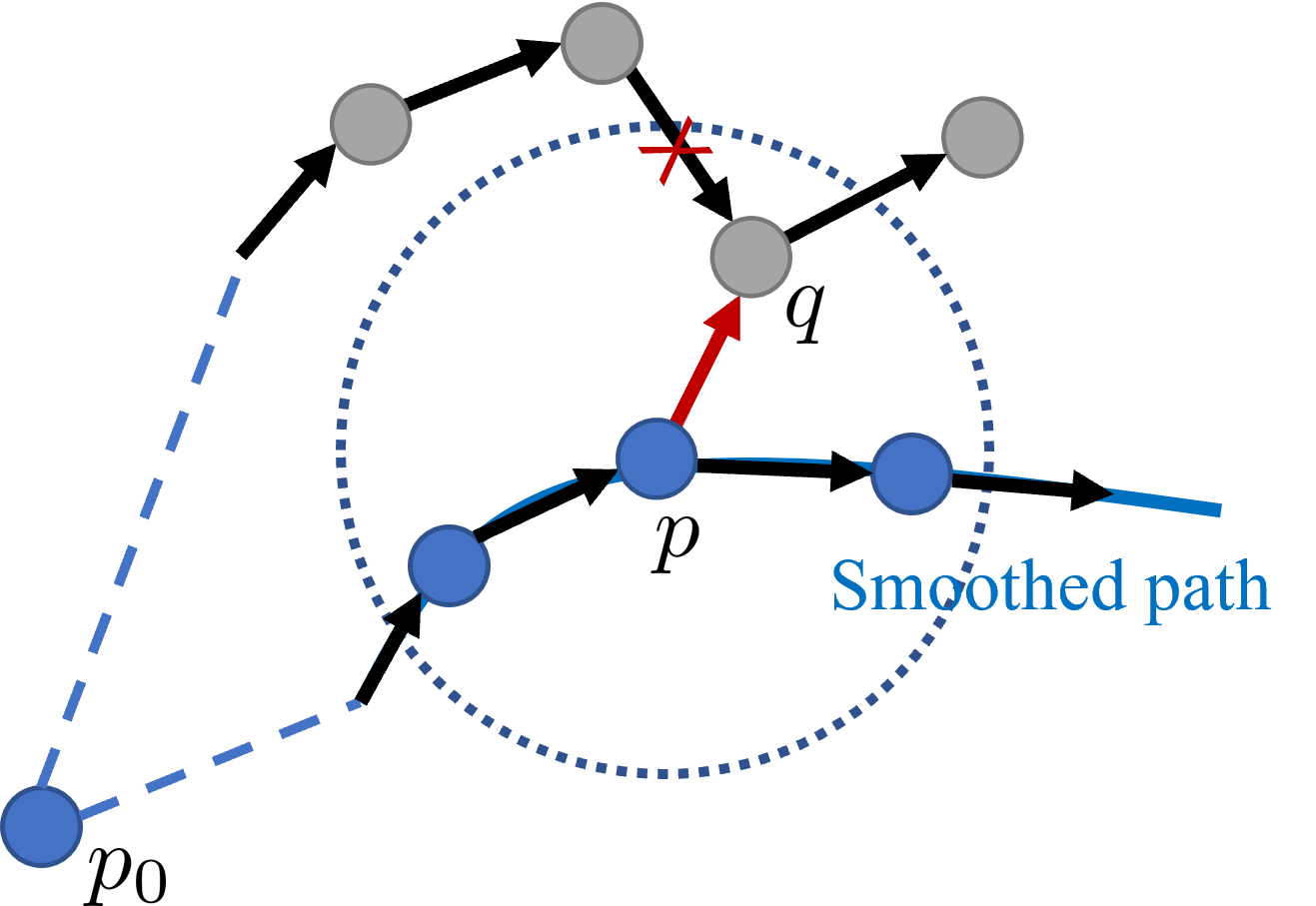}
		\label{fig:forward}}
	\hfil
	\subfloat[]{\includegraphics[width=0.48\linewidth]{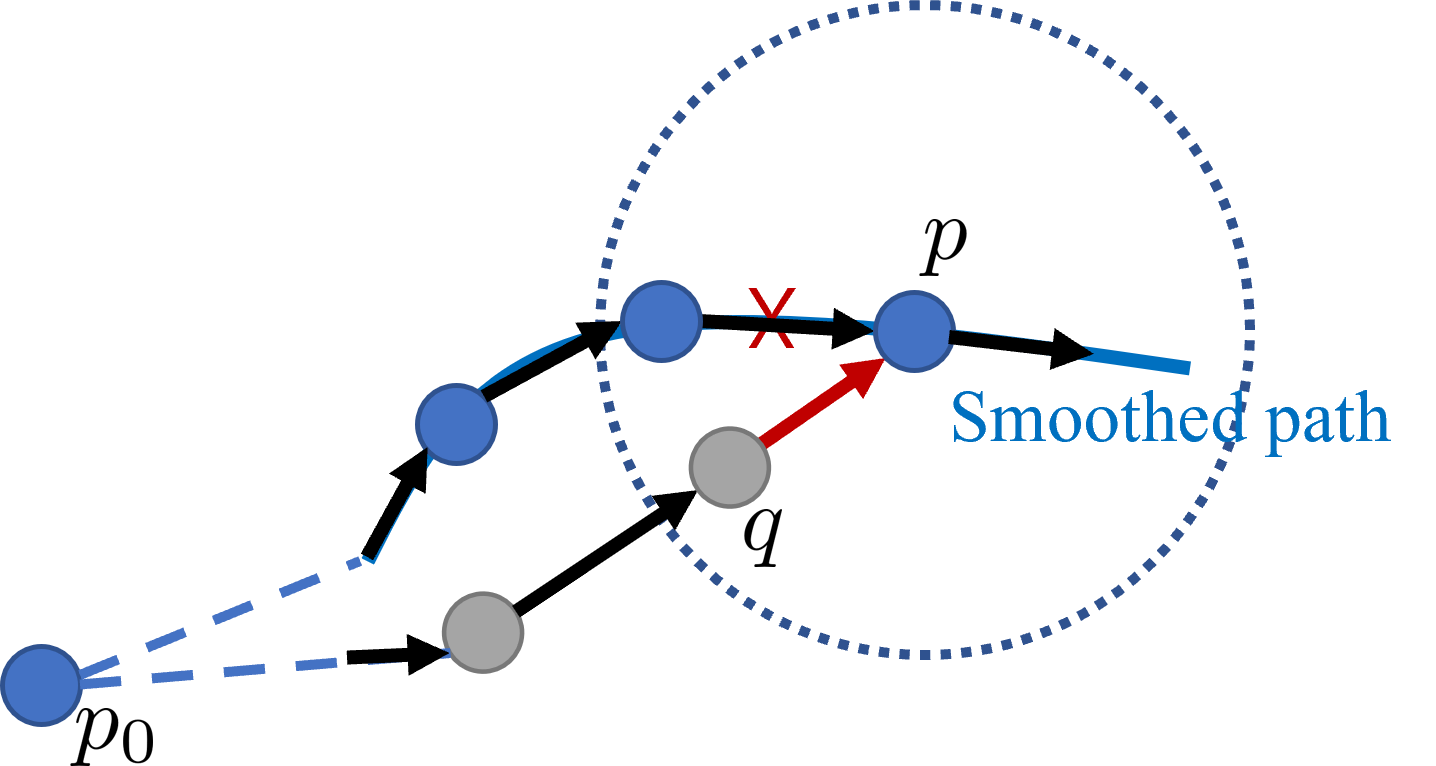}
		\label{fig:reverse}}
	\caption{Illustration of bidirectional rewiring around a smoothed path node \( p \).
		(a) In forward rewiring, the parent of a neighbor \( q \) is updated to \( p \) if it yields a lower cost and the edge is collision-free.
		(b) In reverse rewiring, \( p \) adopts \( q \) as its new parent under similar conditions.
		Solid arrows indicate parent-to-child direction, red arrows show newly rewired edges, and dashed circles represent the rewiring radius.}
	\label{fig:rewiring}
\end{figure}

The algorithm iterates over each point $p \in \mathcal{P}_{\text{smooth}}$ (line~2), and identifies nearby nodes $\mathcal{N}_{\text{near}}$ using a radius-based neighbor search (line~3). For each such node, rewiring is applied in two directions: \emph{forward rewiring} and \emph{reverse rewiring}. These procedures are illustrated in Fig.~\ref{fig:rewiring}.

In the \emph{forward rewiring} phase (lines~4--7), the algorithm checks whether $p$ can provide a lower-cost path to any neighbor $q \in \mathcal{N}_{\text{near}}$. If the cost-to-come to $p$ plus the edge cost $\text{Cost}(p, q)$ is less than the current cost-to-come to $q$, and the edge $(p, q)$ is obstacle-free (line~5), then $p$ becomes the new parent of $q$, and cost values are updated for $q$ and its descendants (lines~6--7).

In the \emph{reverse rewiring} phase (lines~8--11), the algorithm evaluates whether any neighbor $q$ offers a better connection to $p$. If connecting $p$ through $q$ lowers its cost-to-come and the segment $(q, p)$ is collision-free (line~9), then $q$ becomes the new parent of $p$, and cost updates propagate through the affected subtree (lines~10--11). This step allows not only downstream nodes but also those on the smoothed path itself to benefit from improved connections.

By applying both rewiring directions at each point along $\mathcal{P}_{\text{smooth}}$, the algorithm aligns the tree with the geometry of the smoothed path and improves cost consistency. The resulting structure supports more efficient informed optimization in the subsequent phase.

\begin{algorithm}[t]
	\caption{Enhanced SIRRT*}\label{alg:enhanced_sirrt}
	\KwIn{Start $p_{\text{start}}$, Goal $p_{\text{goal}}$, Grid map $I_{\text{grid}}$}
	\KwOut{Final tree $\mathcal{T}$}
	
	\SetKwProg{Fn}{function}{:}{}
	\Fn{\Func{EnhancedSIRRT*}($p_{\text{start}}, p_{\text{goal}}, I_{\text{grid}}$)}
	{
		$I_{\text{skel}} \leftarrow \Func{Skeletonization}(I_{\text{grid}})$\;
		$P_{\text{corner}} \leftarrow \Func{HarrisCornerDetection}(I_{\text{skel}})$\;
		$\mathcal{T} \leftarrow \Func{GenerateInitialTree}(P_{\text{corner}}, p_{\text{start}}, p_{\text{goal}})$\;
		$\mathcal{P}_{\text{init}} \leftarrow \Func{ExtractPath}(\mathcal{T}, p_{\text{start}}, p_{\text{goal}})$\;
		\textcolor{blue}{$\mathcal{P}_{\text{smooth}} \leftarrow \Func{HybridPathSmoothing}(\mathcal{P}_{\text{init}})$\;}
		\textcolor{blue}{$\mathcal{T} \leftarrow \textsc{InsertSmoothedPath}(\mathcal{T}, \mathcal{P}_{\text{smooth}})$\;}		
		\textcolor{blue}{$\mathcal{T} \leftarrow \Func{BidirectionalRewiring}(\mathcal{T}, \mathcal{P}_{\text{smooth}})$\;}
		\For{$i \leftarrow 1$ \KwTo $N$}
		{
			$p_{\text{rand}} \leftarrow \Func{InformedSample}(i, \mathcal{T})$\;
			$p_{\text{near}} \leftarrow \Func{Nearest}(\mathcal{T}, p_{\text{rand}})$\;
			$p_{\text{new}} \leftarrow \Func{Steer}(p_{\text{near}}, p_{\text{rand}})$\;
			\If{$\Func{ObstacleFree}(p_{\text{new}})$}{
				$P_{\text{near}} \leftarrow \Func{Near}(\mathcal{T}, p_{\text{new}})$\;
				$p_{\text{min}} \leftarrow \Func{ChooseParent}(p_{\text{new}}, P_{\text{near}})$\;
				$\mathcal{T} \leftarrow \Func{InsertNode}(p_{\text{min}}, p_{\text{new}}, \mathcal{T})$\;
				$\mathcal{T} \leftarrow \Func{Rewire}(\mathcal{T}, P_{\text{near}}, p_{\text{new}})$\;
			}
		}
		\Return{$\mathcal{T}$}\;
	}
\end{algorithm}

\subsection{Enhanced SIRRT* Using Hybrid Path Smoothing and Bidirectional Rewiring}
\label{sec:enhanced_sirrt}

Algorithm~\ref{alg:enhanced_sirrt} summarizes the complete pipeline of the proposed Enhanced SIRRT* planner. The algorithm consists of two main phases: initial path generation based on grid map skeletonization (lines~2--8), followed by informed optimization using sampling-based RRT* (lines~9--17).

A skeleton of the input grid map $I_{\text{grid}}$ is computed using morphological thinning (line~2), followed by Harris corner detection to extract salient structural features (line~3). These points, along with the start and goal locations, are used to construct an MST $\mathcal{T}$ via Prim's algorithm (line~4). An initial path $\mathcal{P}_{\text{init}}$ is then extracted from the MST by tracing from the goal node back to the start node (line~5).

Next, the initial path and tree are refined by the proposed enhancements (lines~6--8). To improve geometric quality, the path is updated using the \textsc{HybridPathSmoothing} procedure (Algorithm~\ref{alg:hybrid_path_smoothing}, line~6), which performs cubic spline fitting to generate a splined initial path, followed by collision-aware correction to yield a smoothed initial path. This smoothed path $\mathcal{P}_{\text{smooth}}$ is then merged into the initial tree (line~7) by sequentially inserting its nodes while preserving continuity and parent-child relationships.

The tree is subsequently refined using the \textsc{BidirectionalRewiring} procedure (Algorithm~\ref{alg:bidirectional_rewiring}, line~8), which rewires connections between the smoothed path and nearby nodes in both directions. This improves cost propagation and results in a refined initial path embedded in an updated tree structure.

Following tree refinement, the algorithm proceeds with informed sampling-based optimization (lines~9--17). At each iteration, a sample is drawn from the informed ellipsoidal region (line~10), extended toward its nearest neighbor (line~11), and added to the tree if valid (lines~12--17). Parent selection and local rewiring follow the standard RRT* framework, allowing the solution to incrementally improve while preserving asymptotic optimality.

\begin{figure}[]
	\centering
	\subfloat[]{\includegraphics[width=1.0\linewidth]{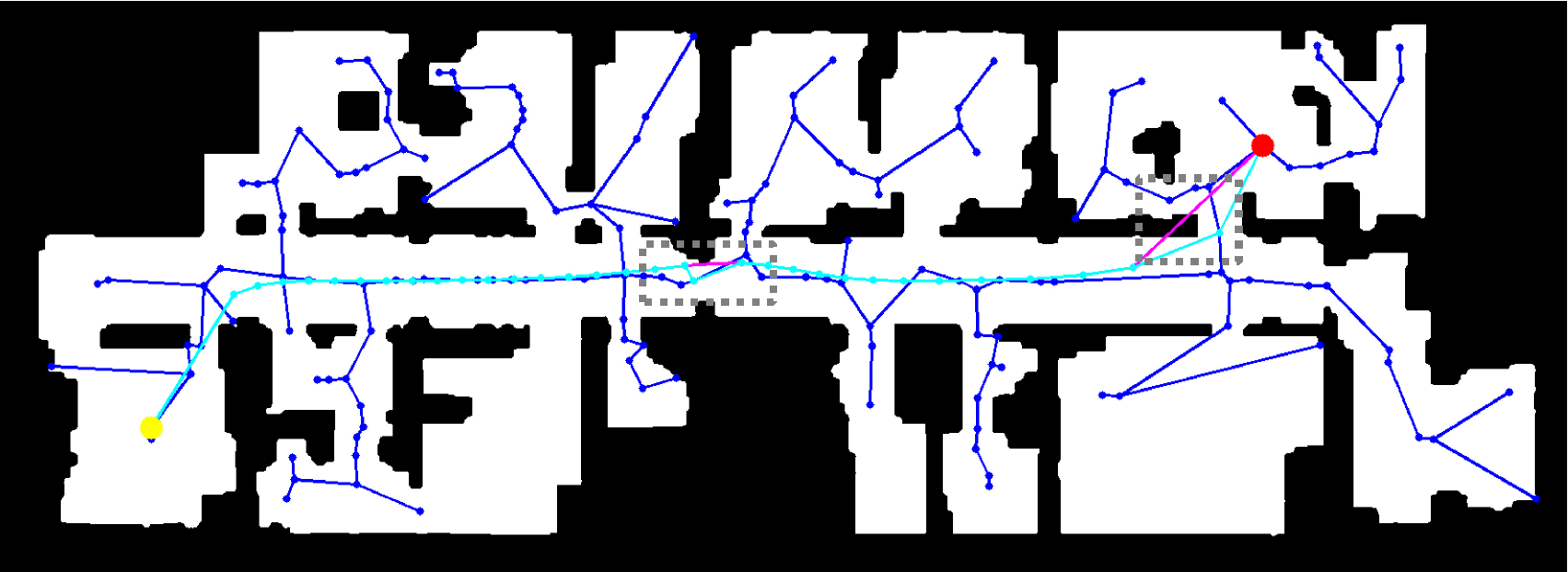}
		\label{fig:tree_merged}}
	\hfil
	\subfloat[]{\includegraphics[width=1.0\linewidth]{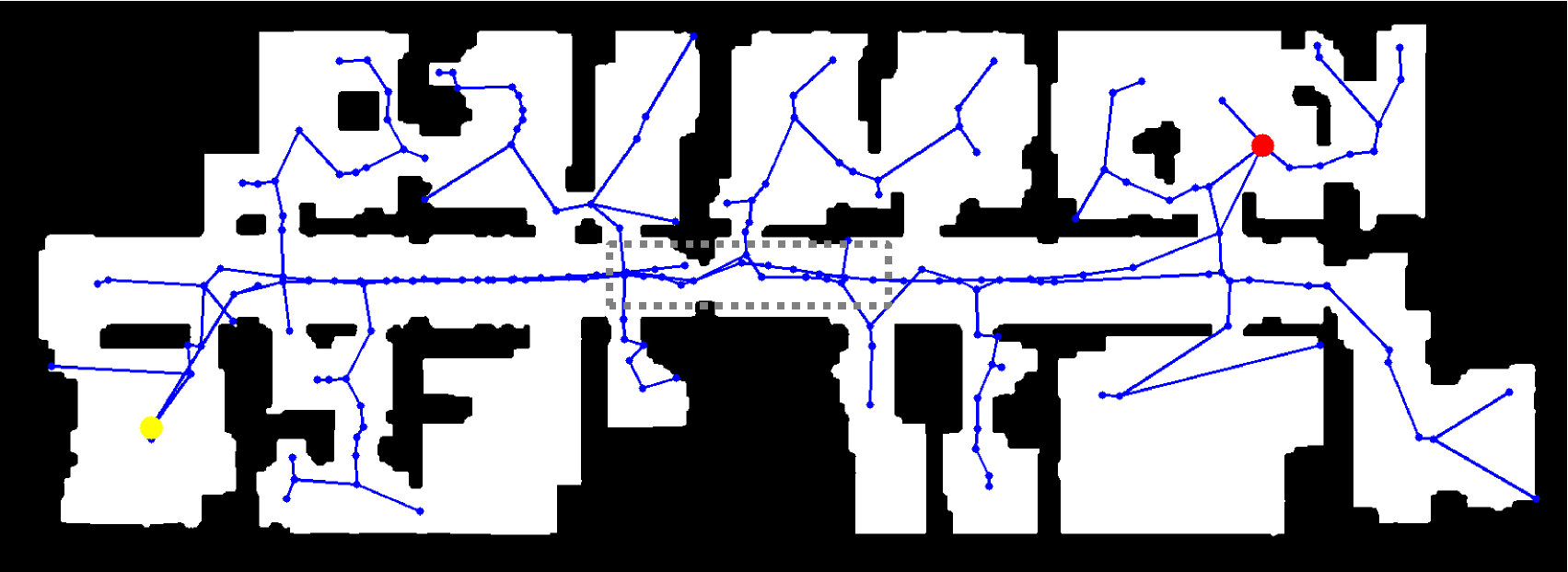}
		\label{fig:tree_rewired_forward}}
	\hfil
	\subfloat[]{\includegraphics[width=1.0\linewidth]{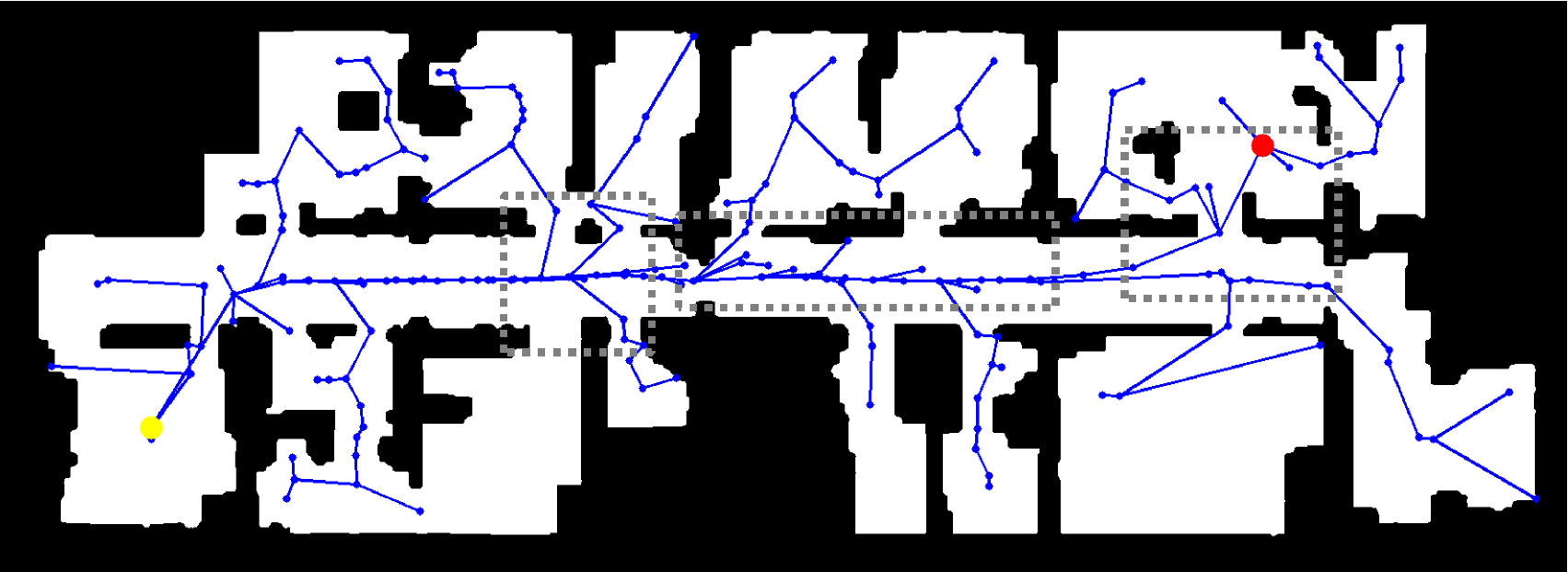}
		\label{fig:tree_rewired_both}}
	\hfil
	\subfloat[]{\includegraphics[width=1.0\linewidth]{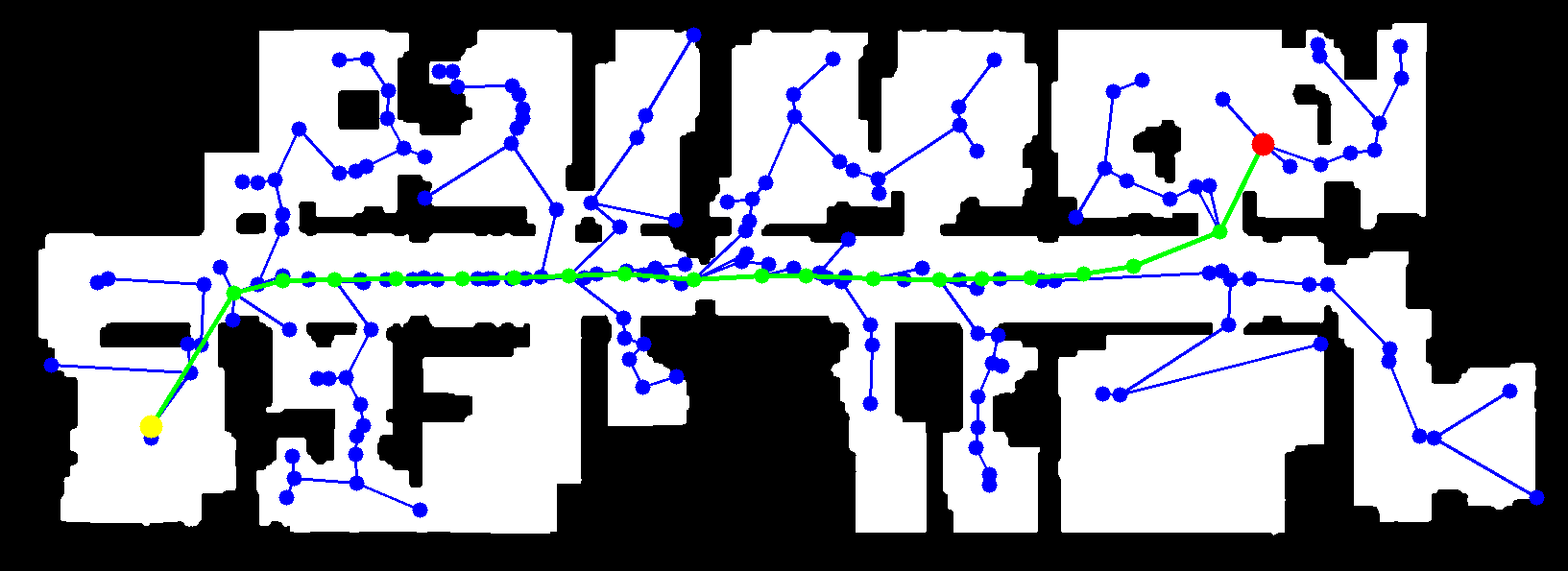}
		\label{fig:tree_final_path}}
	\hfil
	\subfloat[]{\includegraphics[width=1.0\linewidth]{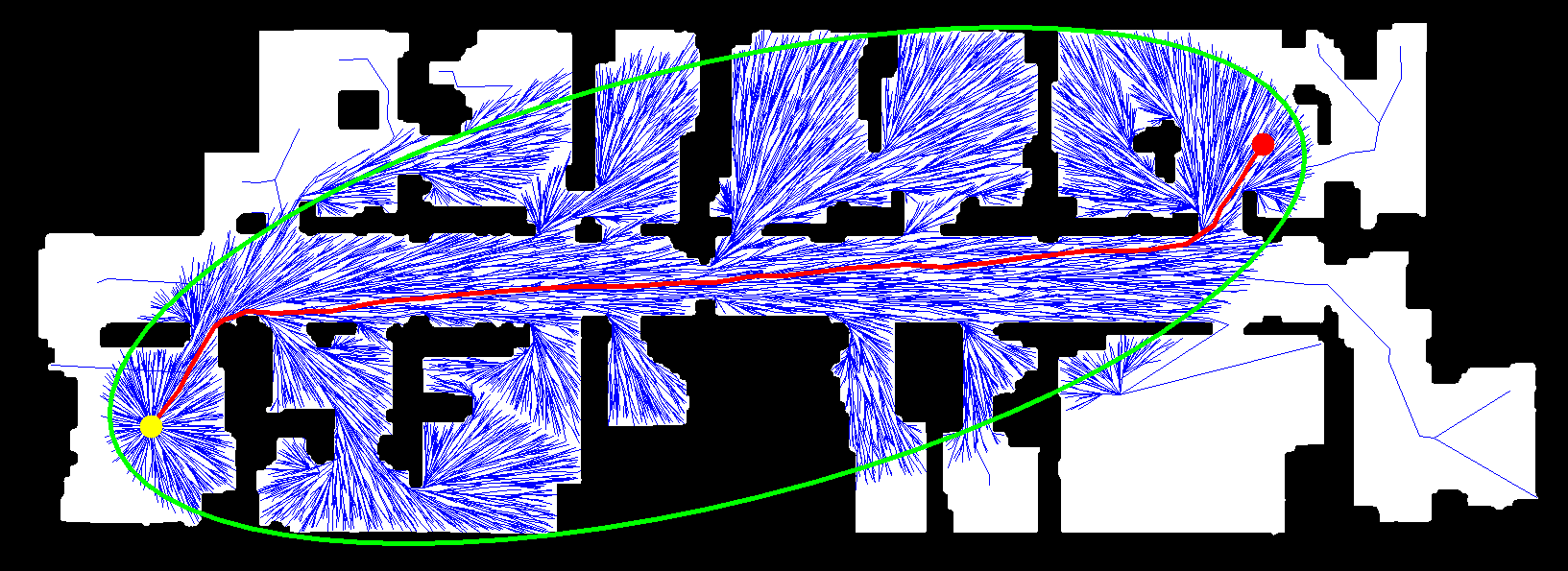}
		\label{fig:result_path}}
	\caption{Initial tree refinement and optimization result in E-SIRRT*.
		(a) Initial tree (blue nodes and edges) constructed via grid map skeletonization, with the smoothed initial path (cyan) merged into the tree. The magenta line shows the splined initial path before collision-aware correction, and corrected segments are highlighted with grey dotted rectangles.
		(b) Tree after forward rewiring around the smoothed path; rewired regions are highlighted with grey dotted rectangles.
		(c) Tree after bidirectional rewiring, further enhancing local connectivity.
		(d) Refined initial tree and path (green).
		(e) Final optimized path (red) obtained through informed sampling within the ellipsoidal sampling region (green ellipse). Blue lines show the expanded tree generated during the optimization phase.
	}\label{fig:esirrt}
\end{figure}

Figure~\ref{fig:esirrt} presents key stages of the E-SIRRT* pipeline. 
Figure~\ref{fig:tree_merged} shows the initial tree constructed from the grid map skeleton, along with the splined initial path (magenta) and the smoothed initial path obtained after collision-aware correction (cyan). 
Figures~\ref{fig:tree_rewired_forward} and~\ref{fig:tree_rewired_both} depict the tree after forward and bidirectional rewiring, respectively, with rewired regions highlighted by dotted rectangles. 
Figure~\ref{fig:tree_final_path} shows the refined initial path (green), extracted from the rewired tree and used as the starting point for informed optimization. 
Finally, Figure~\ref{fig:result_path} presents the final result after informed optimization, including the optimized path (red), the ellipsoidal sampling region (green ellipse), and the expanded tree structure (blue lines).

\section{Experimental Results and Analysis}\label{exp_result} 

\subsection{Experimental Setup}
To evaluate the performance of the proposed E-SIRRT* planner, we conducted experiments in two 2D grid map environments. The first environment, shown in Fig.~\ref{fig:esirrt}, is a modified version of the publicly available benchmark map \emph{Freiburg-079}\cite{mrpt_fr079}, featuring multiple rooms connected by corridors. The second environment, illustrated in Fig.~\ref{fig:exp03}, is a simulated scenario designed specifically to include a narrow passage, which poses a significant challenge to sampling efficiency.

We compared E-SIRRT* against two baselines: the original SIRRT* and IRRT*. Each planner was evaluated over 100 independent trials in both environments. During each trial, we recorded the solution cost at every iteration and analyzed convergence behavior over a fixed number of post-initial iterations. The number of informed optimization iterations after the initial solution was fixed at 20,000 in the first environment and 2,000 in the second. These values were selected to be sufficiently large to allow each planner to approach convergence.

Since IRRT* relies on random sampling to compute its initial solution, the number of iterations required before entering the informed optimization phase varies by trial. Therefore, we also recorded the number of iterations needed to obtain the initial solution in IRRT* as part of the comparative analysis.

All experiments were implemented in C++ and executed on an Intel Core i9-14900K CPU running at 6.0~GHz with 128~GB of RAM. Grid maps were represented as binary occupancy images, and collision checking was performed using OpenCV-based line tracing.

\subsection{Quantitative Comparison}
\begin{table}[]
	\centering
	\caption{Convergence Results Experiment \#1}
	\label{tab:convergence-benchmark_1}
	\renewcommand{\arraystretch}{1.2}
	\begin{tabular}{lccc}
		\toprule
		\makecell{\textbf{}} 
		& \makecell{\textbf{IRRT*}} 
		& \makecell{\textbf{SIRRT*}} 
		& \makecell{\textbf{E-SIRRT*}} \\
		\midrule
		\makecell{Initial \\ Iteration} 
		& \makecell{1922.38 \\ $\pm$ 1291.09 \\ (356--7162)} 
		& \makecell{--}
		& \makecell{--} \\
		\midrule
		\makecell{Initial \\ Cost} 
		& \makecell{1531.91 \\ $\pm$ 106.61 \\ (1391.97--1793.91)} 
		& \makecell{1484.38 \\ $\pm$ 0.00}
		& \makecell{\textbf{1301.84} \\ $\pm$ 0.00} \\
		\midrule
		\makecell{Final \\ Cost} 
		& \makecell{1276.33 \\ $\pm$ 1.76 \\ (1272.85--1283.36)} 
		& \makecell{1276.84 \\ $\pm$ 1.62 \\ (1273.81--1282.24)} 
		& \makecell{\textbf{1276.05} \\ $\pm$ 1.38 \\ (1272.96--1279.49)} \\
		\bottomrule
	\end{tabular}
\end{table}

\begin{figure*}[t]
	\centering
	\subfloat[]{\includegraphics[width=0.32\linewidth]{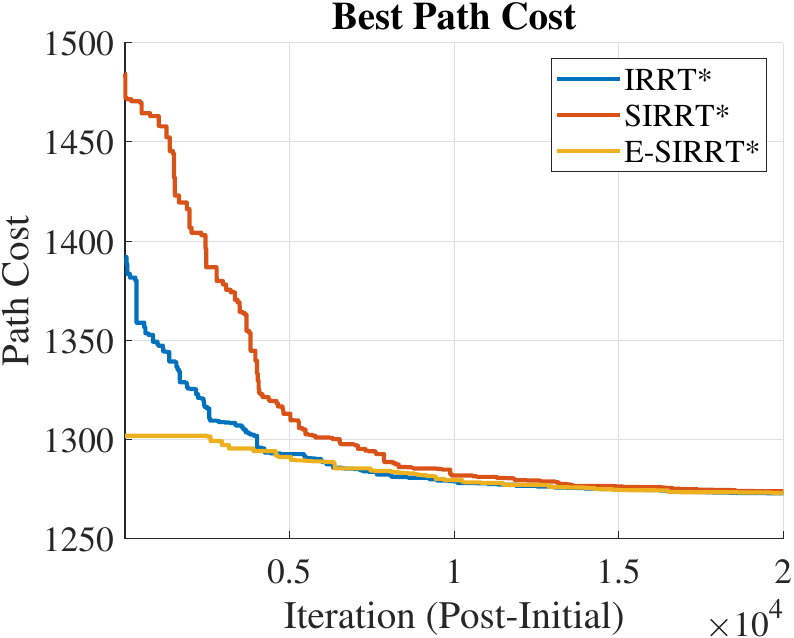}
		\label{fig:best_cost1}}
	\hfil
	\subfloat[]{\includegraphics[width=0.32\linewidth]{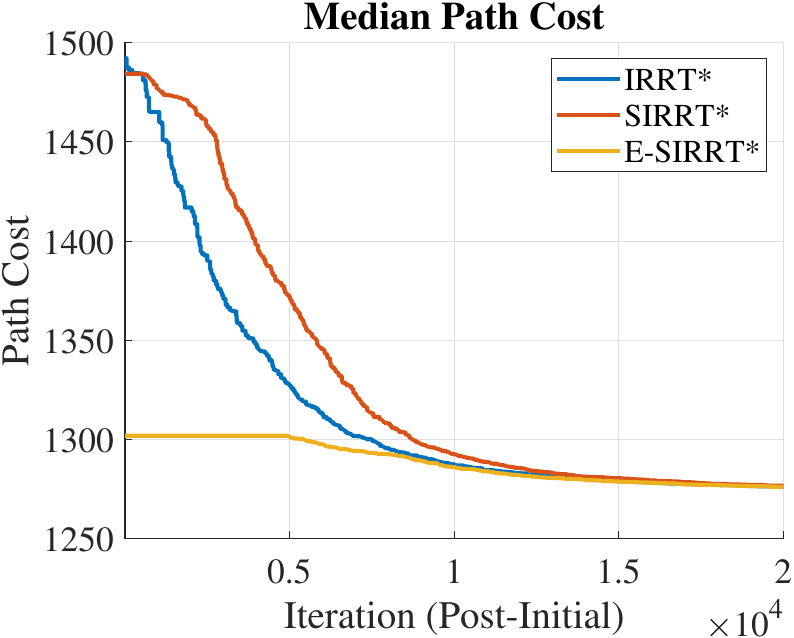}
		\label{fig:median_cost1}}
	\hfil
	\subfloat[]{\includegraphics[width=0.32\linewidth]{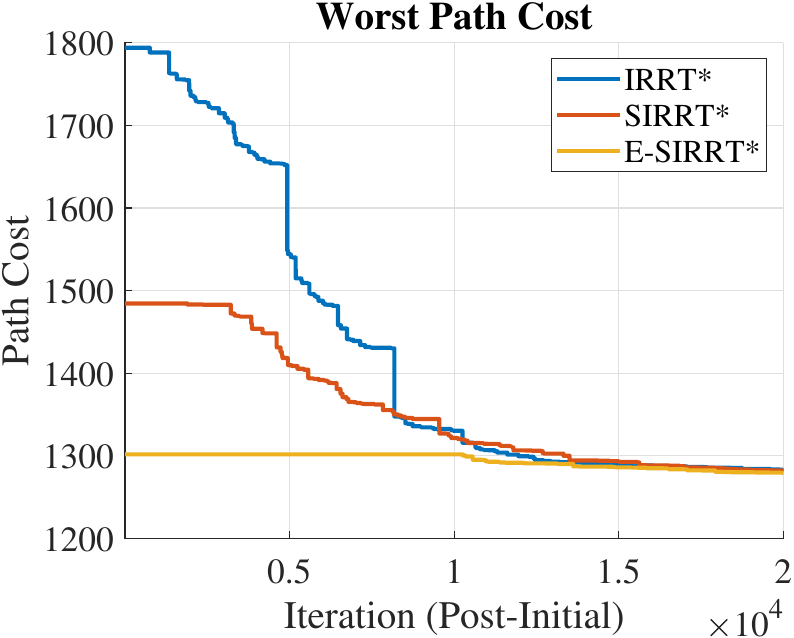}
		\label{fig:worst_cost1}}
	\caption{Convergence of path cost over post-initial iterations for IRRT*, SIRRT*, and Enhanced SIRRT* (E-SIRRT*) in Experiment \#1.
		Each plot shows the path cost after the initial solution is obtained. (a) Best-case performance. (b) Median-case performance. (c) Worst-case performance.}
	\label{fig:cost1}
\end{figure*}

\begin{figure*}[t]
	\centering
	\subfloat[]{\includegraphics[width=0.32\linewidth]{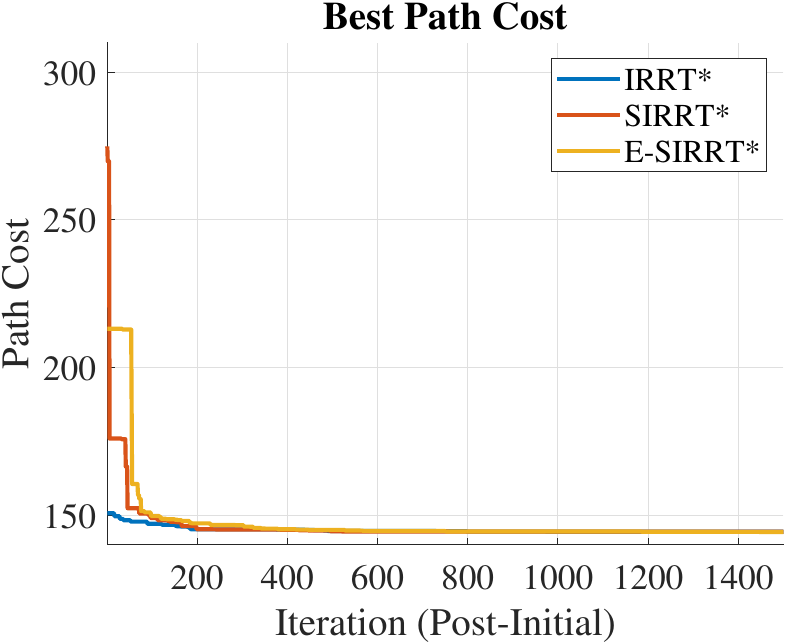}
		\label{fig:best_cost3}}
	\hfil
	\subfloat[]{\includegraphics[width=0.32\linewidth]{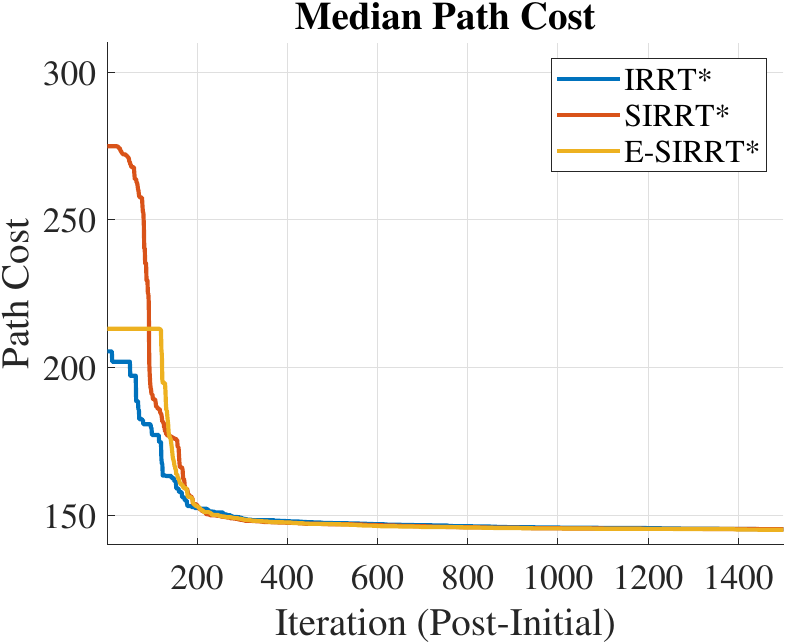}
		\label{fig:median_cost3}}
	\hfil
	\subfloat[]{\includegraphics[width=0.32\linewidth]{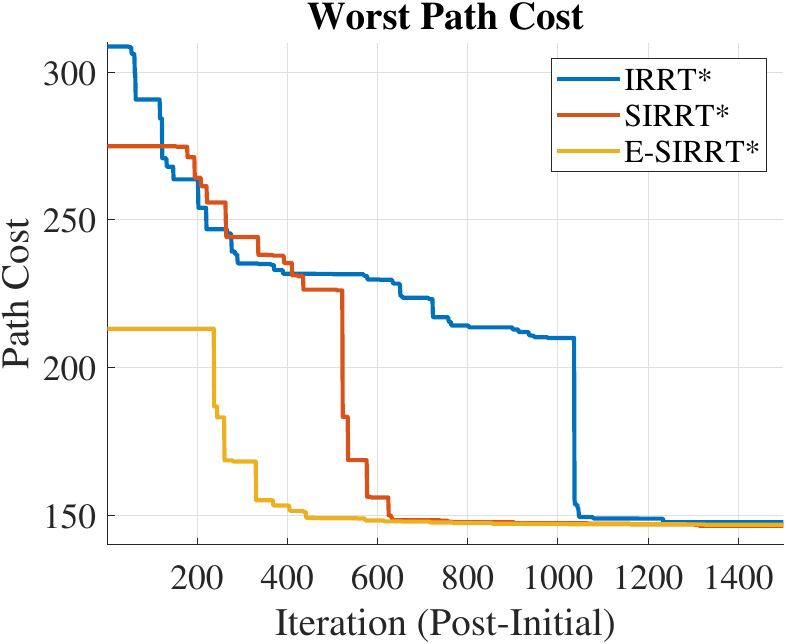}
		\label{fig:worst_cost3}}
	\caption{Convergence of path cost over post-initial iterations for IRRT*, SIRRT*, and Enhanced SIRRT* (E-SIRRT*) in Experiment \#2.
		Each plot shows the path cost after the initial solution is obtained. (a) Best-case performance. (b) Median-case performance. (c) Worst-case performance.}
	\label{fig:cost3}
\end{figure*}

The convergence characteristics of IRRT*, SIRRT*, and the proposed E-SIRRT* are summarized in Tables~\ref{tab:convergence-benchmark_1} and~\ref{tab:convergence-benchmark_3}, and visualized in Figs.\ref{fig:cost1} and \ref{fig:cost3}. Performance is evaluated in terms of initial iteration count (for IRRT*), initial and final path costs, and consistency across 100 independent trials in each environment.

\textbf{Experiment~\#1.} IRRT* exhibits substantial variability in computing an initial solution, with iteration counts ranging from 356 to 7162 (mean: 1922.38, std: 1291.09). This high trial-to-trial variance arises despite identical start and goal positions, highlighting IRRT*'s sensitivity to random sampling. Such unpredictability is particularly problematic in practical applications where robustness and repeatability are essential. In contrast, both SIRRT* and E-SIRRT* deterministically generate initial solutions from the grid map skeleton, resulting in zero variance.

E-SIRRT achieves the best initial path quality, with an initial cost of 1301.84, markedly lower than that of SIRRT* (1484.38) and IRRT* (1531.91). After 20,000 post-initial iterations, E-SIRRT* achieves the lowest final cost (1276.05) with the smallest variance ($\pm$1.38), demonstrating both faster convergence and higher consistency. As shown in Fig. \ref{fig:cost1}, E-SIRRT* consistently outperforms IRRT* and SIRRT* across best-, median-, and worst-case trials. IRRT* converges more slowly and exhibits large fluctuations in the worst case (Fig. \ref{fig:cost1}c), while SIRRT* shows stable but less efficient performance due to the lack of geometric refinement.

\begin{table}[]
	\centering
	\caption{Convergence Results Experiment \#2}
	\label{tab:convergence-benchmark_3}
	\begin{tabular}{lccc}
		\toprule
		\makecell{\textbf{}} 
		& \makecell{\textbf{IRRT*}} 
		& \makecell{\textbf{SIRRT*}} 
		& \makecell{\textbf{E-SIRRT*}} \\
		\midrule
		\makecell{Initial \\ Iteration} 
		& \makecell{118.79 \\ $\pm$ 65.20 \\ (36--497)} 
		& \makecell{--}
		& \makecell{--} \\
		\midrule
		\makecell{Initial \\ Cost} 
		& \makecell{209.11 \\ $\pm$ 43.77 \\ (150.68--308.71)} 
		& \makecell{274.94 \\ $\pm$ 0.00}
		& \makecell{\textbf{213.17} \\ $\pm$ 0.00} \\
		\midrule
		\makecell{Final \\ Cost} 
		& \makecell{145.37 \\ $\pm$ 0.64 \\ (144.41--147.68)} 
		& \makecell{145.25 \\ $\pm$ 0.50 \\ (144.43--146.41)} 
		& \makecell{\textbf{145.23} \\ $\pm$ 0.52 \\ (144.31--146.78)} \\
		\bottomrule
	\end{tabular}
\end{table}

\textbf{Experiment~\#2.} This experiment introduces a narrow passage to evaluate sampling efficiency in constrained settings. IRRT* again shows considerable variation in initial iteration count (mean: 118.79, std: 65.20), reinforcing its unreliability. E-SIRRT* yields an initial cost (213.17) slightly higher than the IRRT* mean (209.11) but with zero variance due to its deterministic initialization. This makes E-SIRRT* significantly more consistent and dependable, especially in early iterations.

Despite similar final costs ($\approx$145) across methods, E-SIRRT* converges with lower variance and smoother progression, as seen in Fig.~\ref{fig:cost3}. IRRT* occasionally achieves competitive performance but suffers from high trial variability. SIRRT* remains consistent but lacks the geometric refinement needed for accelerated convergence.

\begin{figure}[t]
	\centering
	\subfloat[]{\includegraphics[width=0.3\linewidth]{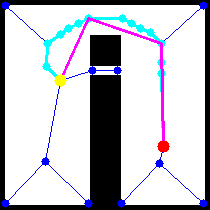}
		\label{fig:esirrt_initial_03}}
	\hfil
	\subfloat[]{\includegraphics[width=0.3\linewidth]{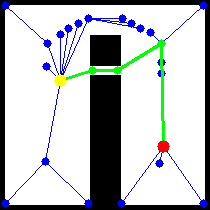}
		\label{fig:tree_rewired_03}}
	\hfil
	\subfloat[]{\includegraphics[width=0.3\linewidth]{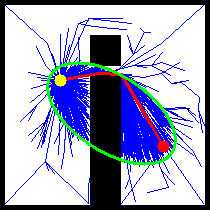}
		\label{fig:esirrt_03}}
	\caption{
		E-SIRRT* result from \textbf{Experiment~\#2}. 
		(a) Initial tree structure (blue nodes and edges) with the extracted MST path (cyan line) and its hybrid-smoothed version (magenta line). 
		(b) Rewired tree structure and the refined initial path (green line) after bidirectional rewiring. 
		(c) Final optimized path (red line) after informed sampling, along with the sampling ellipse (green) used during optimization.}
	\label{fig:exp03}
\end{figure}

A representative result from E-SIRRT* in Experiment~\#2 is shown in Fig. \ref{fig:exp03}. Fig. \ref{fig:esirrt_initial_03} shows the initial tree along with the MST-derived initial path (cyan) and its hybrid-smoothed version (magenta). Although the smoothed path may visually appear less smooth due to collision-aware adjustments, it significantly reduces the path length compared to the initial MST-derived path. Fig.\ref{fig:tree_rewired_03} shows the tree after bidirectional rewiring, where the refined initial path (green) is embedded. Fig.\ref{fig:esirrt_03} presents the final optimized path (red) obtained through informed sampling within the ellipsoidal sampling region.

In summary, these experimental results confirm the motivation of the proposed method: traditional random-sampling-based planners such as IRRT* exhibit high variance and delayed convergence, whereas structure-aware methods provide improved reliability. While SIRRT* achieves robustness through deterministic tree initialization, it lacks geometric refinement capabilities. E-SIRRT* addresses this shortcoming by incorporating hybrid path smoothing and bidirectional rewiring, resulting in high-quality initial paths and trees with consistent cost propagation. 

\section{Conclusion}\label{conclusion}
This paper introduced E-SIRRT*, an advanced structure-aware path-planning algorithm extending the original SIRRT* through hybrid path smoothing and bidirectional rewiring. These enhancements refine the geometry of the initial path and improve tree connectivity, effectively overcoming limitations of existing sampling-based planners related to initialization quality and convergence speed.
Comprehensive experiments conducted in structured and constrained grid-based environments showed that E-SIRRT* consistently delivers faster convergence and more stable performance compared to IRRT* and SIRRT*, while preserving deterministic behavior. The findings validate that combining skeleton-based initialization with principled geometric and structural refinements results in reliable and high-quality motion plans.

Future research will focus on extending E-SIRRT* to higher-dimensional planning scenarios and exploring its integration with task-informed or learning-based sampling strategies to further improve scalability and adaptability.

\bibliographystyle{IEEEtran}
\bibliography{IEEEabrv,esirrt}

\begin{thebibliography}{10}
\providecommand{\url}[1]{#1}
\csname url@rmstyle\endcsname
\providecommand{\newblock}{\relax}
\providecommand{\bibinfo}[2]{#2}
\providecommand\BIBentrySTDinterwordspacing{\spaceskip=0pt\relax}
\providecommand\BIBentryALTinterwordstretchfactor{4}
\providecommand\BIBentryALTinterwordspacing{\spaceskip=\fontdimen2\font plus
\BIBentryALTinterwordstretchfactor\fontdimen3\font minus
  \fontdimen4\font\relax}
\providecommand\BIBforeignlanguage[2]{{%
\expandafter\ifx\csname l@#1\endcsname\relax
\typeout{** WARNING: IEEEtran.bst: No hyphenation pattern has been}%
\typeout{** loaded for the language `#1'. Using the pattern for}%
\typeout{** the default language instead.}%
\else
\language=\csname l@#1\endcsname
\fi
#2}}

\bibitem{karaman2011anytime}
S.~Karaman, M.~R. Walter, A.~Perez, E.~Frazzoli, and S.~Teller, ``Anytime
  motion planning using the rrt,'' in \emph{2011 IEEE international conference
  on robotics and automation}.\hskip 1em plus 0.5em minus 0.4em\relax IEEE,
  2011, pp. 1478--1483.

\bibitem{kuffner2000rrt}
J.~J. Kuffner and S.~M. LaValle, ``Rrt-connect: An efficient approach to
  single-query path planning,'' in \emph{Proceedings 2000 ICRA. Millennium
  conference. IEEE international conference on robotics and automation.
  Symposia proceedings (Cat. No. 00CH37065)}, vol.~2.\hskip 1em plus 0.5em
  minus 0.4em\relax IEEE, 2000, pp. 995--1001.

\bibitem{karaman2011sampling}
S.~Karaman and E.~Frazzoli, ``Sampling-based algorithms for optimal motion
  planning,'' \emph{The international journal of robotics research}, vol.~30,
  no.~7, pp. 846--894, 2011.

\bibitem{gammell2021asymptotically}
J.~D. Gammell and M.~P. Strub, ``Asymptotically optimal sampling-based motion
  planning methods,'' \emph{Annual Review of Control, Robotics, and Autonomous
  Systems}, vol.~4, no.~1, pp. 295--318, 2021.

\bibitem{ryu2019improved}
H.~Ryu and Y.~Park, ``Improved informed rrt* using gridmap skeletonization for
  mobile robot path planning,'' \emph{International Journal of Precision
  Engineering and Manufacturing}, vol.~20, no.~11, pp. 2033--2039, 2019.

\bibitem{gammell2014informed}
J.~D. Gammell, S.~S. Srinivasa, and T.~D. Barfoot, ``Informed rrt*: Optimal
  sampling-based path planning focused via direct sampling of an admissible
  ellipsoidal heuristic,'' in \emph{IEEE/RSJ International Conference on
  Intelligent Robots and Systems (IROS)}, 2014, pp. 2997--3004.

\bibitem{mashayekhi2020informed}
R.~Mashayekhi, M.~Y.~I. Idris, M.~H. Anisi, I.~Ahmedy, and I.~Ali, ``Informed
  rrt*-connect: An asymptotically optimal single-query path planning method,''
  \emph{IEEE Access}, vol.~8, pp. 19\,842--19\,852, 2020.

\bibitem{klemm2015rrt}
S.~Klemm, J.~Oberl{\"a}nder, A.~Hermann, A.~Roennau, T.~Schamm, J.~M. Zollner,
  and R.~Dillmann, ``Rrt*-connect: Faster, asymptotically optimal motion
  planning,'' in \emph{2015 IEEE international conference on robotics and
  biomimetics (ROBIO)}.\hskip 1em plus 0.5em minus 0.4em\relax IEEE, 2015, pp.
  1670--1677.

\bibitem{wang2024improved}
Y.~Wang, C.~Liu, J.~Li, Q.~Tang, and Y.~Yang, ``An improved informed
  rrt*-connect algorithm for uavs path planning,'' in \emph{IEEE International
  Conference on Unmanned Systems (ICUS)}, 2024.

\bibitem{huang2024neural}
Z.~Huang, H.~Chen, J.~Pohovey, and K.~Driggs-Campbell, ``Neural informed rrt*:
  Learning-based path planning with point cloud state representations under
  admissible ellipsoidal constraints,'' in \emph{2024 IEEE International
  Conference on Robotics and Automation (ICRA)}.\hskip 1em plus 0.5em minus
  0.4em\relax IEEE, 2024, pp. 8742--8748.

\bibitem{chi2018risk}
W.~Chi, J.~Wang, and M.~Q.-H. Meng, ``Risk-informed-rrt*: A sampling-based
  human-friendly motion planning algorithm for mobile service robots in indoor
  environments,'' in \emph{2018 IEEE international conference on information
  and automation (ICIA)}.\hskip 1em plus 0.5em minus 0.4em\relax IEEE, 2018,
  pp. 1101--1106.

\bibitem{chen2024rbi}
F.~Chen, Y.~Zheng, Z.~Wang, W.~Chi, and S.~Liu, ``Rbi-rrt*: Efficient
  sampling-based path planning for high-dimensional state space,'' in
  \emph{2024 IEEE International Conference on Robotics and Automation
  (ICRA)}.\hskip 1em plus 0.5em minus 0.4em\relax IEEE, 2024, pp. 8721--8727.

\bibitem{armstrong2021rrt}
D.~Armstrong and A.~Jonasson, ``Am-rrt*: Informed sampling-based planning with
  assisting metric,'' in \emph{2021 IEEE International Conference on Robotics
  and Automation (ICRA)}.\hskip 1em plus 0.5em minus 0.4em\relax IEEE, 2021,
  pp. 10\,093--10\,099.

\bibitem{gammell2020batch}
J.~D. Gammell, T.~D. Barfoot, and S.~S. Srinivasa, ``Batch informed trees
  (bit*): Informed asymptotically optimal anytime search,'' \emph{The
  International Journal of Robotics Research}, vol.~39, no.~5, pp. 543--567,
  2020.

\bibitem{strub2020advanced}
M.~P. Strub and J.~D. Gammell, ``Advanced bit*(abit*): Sampling-based planning
  with advanced graph-search techniques,'' in \emph{2020 IEEE International
  Conference on Robotics and Automation (ICRA)}.\hskip 1em plus 0.5em minus
  0.4em\relax IEEE, 2020, pp. 130--136.

\bibitem{cormen2009introduction}
T.~H. Cormen, C.~E. Leiserson, R.~L. Rivest, and C.~Stein, \emph{Introduction
  to Algorithms}, 3rd~ed.\hskip 1em plus 0.5em minus 0.4em\relax MIT Press,
  2009.

\bibitem{quarteroni2007scientific}
A.~Quarteroni, F.~Saleri, and P.~Gervasio, \emph{Scientific Computing with
  MATLAB and Octave}, 3rd~ed., ser. Texts in Computational Science and
  Engineering.\hskip 1em plus 0.5em minus 0.4em\relax Springer, 2007, vol.~2.

\bibitem{mrpt_fr079}
{MRPT Project}, ``{Freiburg-079 Building Corridor Dataset},''
  \url{https://www.mrpt.org/Dataset_fr079}, 2020, accessed: 2024-05-24.

\end{thebibliography}
\end{document}